\documentclass{article}

\PassOptionsToPackage{numbers, compress}{natbib}

\usepackage[preprint]{neurips_2025}

\usepackage[utf8]{inputenc} 
\usepackage[T1]{fontenc}    
\usepackage{hyperref}       
\usepackage{url}            
\usepackage{booktabs}       
\usepackage{amsfonts}       
\usepackage{nicefrac}       
\usepackage{microtype}      
\usepackage{xcolor}         

\definecolor{systemcolor}{HTML}{F4A261}
\definecolor{usercolor}{HTML}{2A9D8F}

\usepackage{amssymb}
\usepackage{amsmath}
\usepackage[inline]{enumitem}
\usepackage{wrapfig}
\usepackage{multirow}
\usepackage{textcomp}
\usepackage{subfigure}
\usepackage{caption}
\usepackage{longtable}
\usepackage{fdsymbol}
\usepackage{cleveref}

\crefname{section}{\S}{\S\S}
\Crefname{section}{\S}{\S\S}
\crefname{appendix}{\S}{\S\S}
\Crefname{appendix}{\S}{\S\S}
\crefname{figure}{Figure}{Figures}
\Crefname{figure}{Figure}{Figures}
\crefname{table}{Table}{Tables}
\Crefname{table}{Table}{Tables}
\crefname{equation}{Eq.}{Eqs.}
\Crefname{equation}{Eq.}{Eqs.}

\usepackage[disable]{todonotes}

\newcommand{\ProgRM}{\textsc{ProgRM}}
\newcommand{\progrm}{\textsc{ProgRM}}

\newcommand{\recipe}{recipe}

\newif\ifwithsupp
\withsupptrue

\title{\ProgRM{}: Build Better GUI Agents with Progress Rewards}

\author{%
	Danyang Zhang$^{1,\:\!2}$\:\!\thanks{Equal contribution.} \quad
	Situo Zhang$^{1,\:\!2\:\!\dagger}$ \quad
	Ziyue Yang$^{1,\:\!2}$ \quad
	Zichen Zhu$^{1,\:\!2}$ \\
	\textbf{Zihan Zhao}$^{1,\:\!2}$ \quad
	\textbf{Ruisheng Cao}$^{1,\:\!2}$ \quad
	\textbf{Lu Chen}$^{1,\:\!2,\:\!3}$\:\!\thanks{Corresponding authors.} \quad
	\textbf{Kai Yu}$^{1,\:\!2,\:\!3\:\!\ddagger}$ \\
	$^{1}$X-LANCE Lab, School of Computer Science \\
	MoE Key Lab of Artificial Intelligence, SJTU AI Institute \\
	Shanghai Jiao Tong University, Shanghai, China \\
	$^{2}$Jiangsu Key Lab of Language Computing, Suzhou, China \\
	$^{3}$Suzhou Laboratory, Suzhou, China \\
	\texttt{\{zhang-dy20,situozhang\}@sjtu.edu.cn} \\
}

\begin{document}

\setcounter{footnote}{1}
\setcounter{Hfootnote}{1}
\maketitle
\setcounter{footnote}{0}
\setcounter{Hfootnote}{0}

\begin{abstract}
  LLM-based (Large Language Model) GUI (Graphical User Interface) agents can potentially reshape
  our daily lives significantly. However, current LLM-based GUI agents suffer from the scarcity
  of high-quality training data owing to the difficulties of trajectory collection and reward annotation.
  Existing works have been exploring LLMs to collect trajectories for imitation learning or to
  offer reward signals for online RL training. However, the Outcome Reward Model (ORM) used in existing
  works cannot provide finegrained feedback and can over-penalize the valuable steps in finally failed
  trajectories. To this end, we propose \textbf{Prog}ress \textbf{R}eward \textbf{M}odel (\ProgRM{})
  to provide dense informative intermediate rewards by predicting a task completion progress
  for each step in online training. To handle the challenge of progress reward label annotation,
  we further design an efficient LCS-based (Longest Common Subsequence) self-annotation algorithm to
  discover the key steps in trajectories and assign progress labels accordingly. \ProgRM{} is evaluated
  with extensive experiments and analyses. Actors trained with \progrm{} outperform leading proprietary
  LLMs and ORM-trained actors, illustrating the effectiveness of \progrm{}. The codes for experiments
  will be made publicly available upon acceptance.
\end{abstract}

\section{Introduction}


Automatic Graphical User Interface (GUI) agents have great potential to reshape our daily lives by automating 
routine operations on GUI systems like computers and smartphones.
Recently, surprised by the exceptional achievements of Large Language Models (LLM) in common-sense and reasoning domains, 
LLMs have been explored to improve the performance of GUI agents~\citep{Rememberer,AppAgent,MobileAgent,UFO,SeeAct,SteP,MobA,DigiRL,WebRL,Aguvis,UI-TARS,AgentTrek,Synatra,EEF,UIR1,GUIR1}.

Using off-the-shelf LLMs to perform GUI tasks through prompt-based methods often yields unsatisfactory results, as these models lack the ability to ground instructions to low-level actions or to make long-term decisions required for GUI tasks. While there has been a growing body of work on training LLMs to build GUI agents, such training typically relies on human-labeled tasks or trajectories. However, annotating GUI tasks is labor-intensive, demands domain-specific expertise, and is extremely difficult to scale. The scarcity of high-quality training data presents a major challenge in developing high-performance GUI agents.



\begin{figure}[t]
    \centering
    \includegraphics[width=0.9\linewidth]{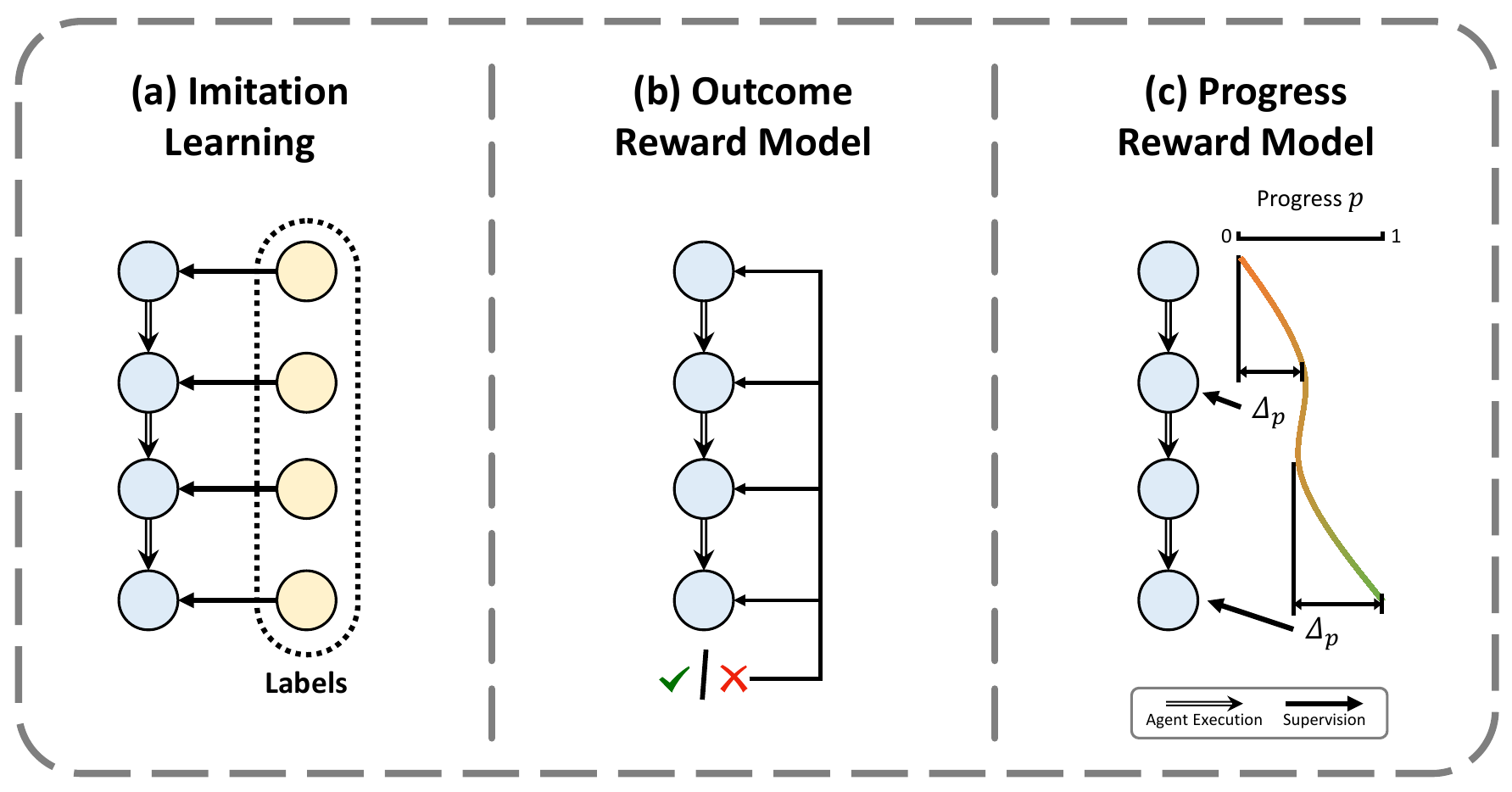}  
    \caption{Comparison of policy optimization methods. (a) Imitation Learning optimizes the agent’s policy using per-step expert labels. (b) ORM provides sparse rewards by updating the policy only based on the trajectory’s final success or failure. (c) \ProgRM{} predicts progress value at each step, using the progress gain ($\Delta_p$) as dense reward signal.}
    \label{fig:compare}
    \vspace{-0.5cm}
\end{figure}

To address this challenge, many studies have explored ways to automatically
synthesize GUI training data~\citep{PAE, LearnByInteract, Explorer}. These
approaches primarily leverage LLMs to generate new task instructions and
collect trajectories produced by actor LLMs. NNetnav~\citep{NNetnav} and
OS-Genesis~\citep{OSgenesis} conclude task instructions after collecting
trajectories, while Explorer~\citep{Explorer} further iterates on this process.
However, these methods rely on imitation learning (see \cref{fig:compare}(a))
and are not well-suited for online environments, where content changes
dynamically and agents may encounter unseen scenarios. They lack the ability to
learn from mistakes or benefit from online exploration to improve performance.

Efficient online Reinforcement Learning (RL) requires reliable reward signals to accurately evaluate arbitrary GUI-based tasks. DistRL~\citep{DistRL} employs an additional Vision-Language Model (VLM) as an autonomous evaluator to determine task success, while WebRL~\citep{WebRL} trains a lightweight Outcome Reward Model (ORM). However, these methods rely solely on the final success status, unnecessarily penalizing all steps within trajectories that fail to achieve the goal but potentially include valuable intermediate actions \citep{EEF}, as shown in \cref{fig:compare}(b). Additionally, the sparse reward signals from ORM significantly reduce exploration efficiency in RL training, particularly for long-horizon tasks typical in GUI interactions.

To this end, we introduce the \textbf{Prog}ress \textbf{R}eward \textbf{M}odel (\ProgRM{}), an effective and efficient approach for generating reward signals at intermediate steps within RL training trajectories for GUI agents. Intuitively, a complex task is not accomplished in a single leap; instead, it is completed through a sequence of subtasks that progressively advance toward the final goal. As shown in \Cref{fig:compare}(c), at each stage, we can quantify how much of the overall task has been completed --- this is the essence of \emph{progress}. For example, when booking a flight, the process typically involves searching for available flights, selecting a suitable option, entering passenger details, and completing the payment. Each of these steps brings the agent closer to the goal, and \progrm{} provides informative reward signals by estimating the progress achieved at each step, rather than waiting until the entire task is finished. This dense, intermediate reward signal enables more efficient and stable RL training, especially for long-horizon tasks commonly encountered in GUI environments.

However, obtaining accurate progress reward labels poses a significant challenge for training reliable \progrm{} models,
as gold labels are commonly not directly available in raw trajectories. This problem is conceptually similar to
the difficulties of process reward annotation faced in Process Reward Model (PRM) training in reasoning tasks.
Human experts or Monte-Carlo search are commonly employed in the reasoning area to label
rewards for intermediate reasoning steps~\citep{VerifyStepByStep, MathShepherd, zhang2025lessons}. However, such methods are prohibitively expensive and time-consuming for GUI agent tasks, where heavy simulators are involved and rapid rollback or efficient state restoration is often unsupported. 
Therefore, to address the progress reward labeling challenge, we propose an efficient self-annotation algorithm that automatically identifies {\em key steps} within trajectories and assigns progress labels accordingly. Specifically, for a given task, we first
discover the common patterns from the successful trajectories by computing their Longest Common Subsequences (LCS)
and extract them as execution \emph{\recipe{s}}. The \recipe{s}
are then used to identify the key steps in unseen trajectories, and the progress labels can be efficiently assigned based on the
identified key steps. The proposed progress labeling algorithm prevents expensive human-expert annotation and Monte-Carlo search,
while it can easily and sufficiently exploit the self-explored trajectories of agents.

We evaluate the effectiveness of \progrm{} on WikiHow task set~\citep{MobileEnv}, a real-world Android device navigation benchmark. Experimental results demonstrate that \progrm{-trained} actors outperform leading proprietary LLMs for GUI tasks, including Claude-3.7-Sonnet, achieving significant superiority in success rate. It also surpasses existing imitation learning and ORM-based online RL approaches. Furthermore, we show that \progrm{} accurately captures the progresses agents make during navigation, highlighting its capability of providing meaningful intermediate feedback in complex environments.

The key contributions of our work are as follows:
\begin{itemize}[leftmargin=12pt]
\item We propose \progrm{}, a novel method that provides dense reward signals based on predicted progresses toward a goal, enabling effective online RL training for GUI agents.
\item We introduce an efficient LCS-based self-annotation algorithm to automatically generate progress labels for training \progrm{}.
\item Experimental results on real-world GUI benchmark demonstrate the superiority of \progrm{-trained} actors to state-of-the-art proprietary models as well as imitation-learning and ORM-training agents.
\end{itemize}

\section{\ProgRM{}: Progress reward model for GUI RL}
\label{sec:method}

\subsection{Progress reward}
\label{sub:prog_rwd_def}

\emph{Progress} is defined as the percentage of a task that has been completed. We introduce the \emph{progress function} $\mathtt{Prog}\in[0,1]$, which estimates the progress $p_t$ given a state $s_t$ during the execution of a task $g$:
\begin{equation}
	p_t = \mathtt{Prog}(s_t; g).
	\label{eqn:prog_func}
\end{equation}
Based on the progress function $\mathtt{Prog}$, we define the progress reward $r^{(p)}_t$ for state $s_t$ as:
\begin{equation}
	r^{(p)}_t = \mathtt{Prog}(s_t; g) - \mathtt{Prog}(s_{t-k}; g),
	\label{eqn:progrm_def}
\end{equation}
where $k$ is a hyperparameter that determines the length of the progress history. 
The progress reward awards the agent with the cumulative progress gain over the past $k$ steps.

\begin{figure}[t]
	\centering
	\includegraphics[width=0.9\linewidth]{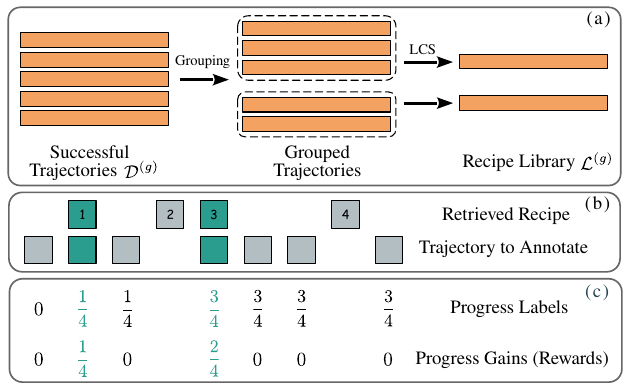}
	\caption{Progress labeling algorithm. The proposed labeling algorithm
		consists of three stages: (a) LCS \recipe{} library construction, where
		trajectories sharing similar core policies are grouped and the common
		pattern called \emph{\recipe{}} within each group is extracted by
		computing the group LCS; (b) Key Step Discovery, in which key steps
		(matched steps, highlighted in green; best viewed in color) are
		identified by matching each trajectory to the \recipe{} with the
		highest completion ratio (see \cref{eqn:lcs_sim}); and (c) Progress
	Label Assignment, where progress labels for key steps are determined by
their position within the \recipe{}, and labels for non-key steps are inherited
from their nearest preceding key step. Once progress values are assigned,
per-step progress gains are used as rewards.}
	\label{fig:prg_lbl}
\end{figure}

\subsection{Progress labeling}\label{sec:process_label}

Since the true progress values are not directly observable from raw interaction trajectories, an appropriate estimation method is required. A naive approach is to assign linear progress label to each step in a successful trajectory, for example, given a trajectory of length $T$, the progress at the $t$-th step is labeled as $p_t^\ast = t / T$. However, this approach assumes uniform progress gain throughout the trajectory, which is often not the case. The actor may take some exploration or useless
actions that do not result in actual new progress towards task completion. To enable reliable
progress estimation, such ``hollow'' steps need to be distinguished
from the steps that actually make progress gains towards the final goal, which we call {\em key steps}.
Besides, actors do not always fail at the exact episode beginning and can make partial progress even 
in a finally failed trajectory. In such cases, the naive linear progress labeling cannot provide meaningful labels
and may cause underestimation of the task progress.

To address these limitations, we propose an algorithm that automatically identifies key steps from successful trajectories and generates more refined progress labels accordingly. As shown in \cref{fig:prg_lbl}, the proposed progress labeling algorithm consists of three stages:
\begin{enumerate*}[label=(\arabic*)]
\item Longest Common Subsequence (LCS) \recipe{} library construction,
\item key step discovery,
\item and progress label assignment.
\end{enumerate*}

\paragraph{LCS \recipe{} library construction}
A natural assumption for key step discovery is that the successful trajectories for the same task goal 
share some common behavior patterns.
Therefore, the common parts of successful trajectories are more likely to be the key steps. From such a perspective,
we propose to extract the common parts of the successful trajectories and store them as \emph{\recipe{s}} and use
the stored \recipe{s} to discover the key steps in unseen trajectories.

Specifically, we start by collecting all successful trajectories for task goal $g$, denoted 
as $\mathcal{D}^{(g)} = \{T_1, T_2, \cdots, T_n\}$. Since there is often more than one optimal policy for completing a given task, we group the trajectories and assume that those within the same group share a common core policy. We then extract one \recipe{} from each group.
Grouping is performed according to LCS-based similarity, ensuring that the similarity between any two trajectories
within a group exceeds a predefined threshold $\theta_L$. \todo[disable]{The value of $\theta_L$ is not mentioned in the paper.}
The similarity between two trajectories is defined as:
\begin{equation}
\mathtt{Sim}(T_i, T_j) = \dfrac{\mathtt{SoftLCS}(T_i, T_j)}{\min\{|T_i|, |T_j|\}},
\label{eqn:slcs_sim}
\end{equation}
where $\mathtt{SoftLCS}$ denotes the customized soft LCS length between trajectories $T_i$ and $T_j$. 
The proposed soft LCS algorithm replaces the exact matching in the traditional LCS algorithm with a soft match function, allowing different types of actions to be matched with varying weights. This enables the algorithm to more effectively handle actions that include natural language arguments, such as text typing. The detailed definition of the soft LCS function is provided in \cref{sub:slcs_def}.

After grouping, we extract \recipe{} for each group by computing the trajectories' LCS and
attain a \recipe{} library $\mathcal{L}^{(g)} = \{L_1, L_2, \cdots, L_m\}$. 

\paragraph{Key step discovery}
The second stage involves identifying the key steps within a given trajectory $T_i$ for task $g$, using the constructed LCS
\recipe{} library $\mathcal{L}^{(g)}$. For each trajectory, we first select the \recipe{} $L_j \in \mathcal{L}^{(g)}$ that maximizes the completion ratio, \textit{i.e.}, the proportion of the \recipe{} matched with the trajectory to annotate.
Denoting the LCS between $T_i$ and $L_j$ as $l_{ij}$, the completion ratio ($\mathtt{CR}$) is formulated as
\begin{equation}
\mathtt{CR}(T_i; L_j) = \dfrac{|l_{ij}|}{|L_j|}.
\label{eqn:lcs_sim}
\end{equation}
Here, $|l_{ij}|$ denotes the length of the LCS, and $|L_j|$ is the length of the \recipe{}. The steps in $T_i$ that also appear in $L_j$ are regarded as key steps, representing critical progress milestones within the trajectory.

\paragraph{Progress label assignment}
Progress labels are then assigned separately for key steps and non-key steps. For each key step, its progress label is determined by its position within the matched \recipe{} $L_j$, under the assumption that progress increases uniformly along the \recipe{}. Specifically, if the $\lambda$\mbox{-}th key step in $T_i$ corresponds to the $\kappa$\mbox{-}th position in $L_j$, its progress label is given by $p_{k_\lambda}^\ast = \kappa / |L_j|$. For non-key steps (\textit{i.e.}, steps between two key steps), we assign the progress label of their nearest preceding key step, \textit{i.e.}, for a non-key step $k_\lambda < t < k_{\lambda+1}$, its progress label is $p_t^\ast = p_{k_\lambda}^\ast$.

For environments that provide milestone-style intermediate rewards, these rewards can be used directly to identify key steps. In this case, key steps correspond to those receiving milestone rewards, and progress labels can be assigned using the same approach as above.

\subsection{Progress model training}

\todo[disable]{Model structure and loss}

To develop a practical progress model, we combine a pretrained LLM with a multilayer perceptron (MLP) and apply a sigmoid activation to ensure the output is constrained between 0 and 1. The model is trained using the binary cross-entropy (BCE) loss, which is well-suited for optimizing normalized progress value predictions.
Given a training dataset of interaction steps and their corresponding progress labels, $\mathcal{D}^{(p)} = \{(g_i, s_i, p_i^\ast)\}$, the progress model parameterized by $\theta$ is optimized as follows:
\begin{equation}
\begin{aligned}
\hat{p}_i &= \mathtt{Prog}([g_i, \hat{s}_i]; \theta), \\
\mathcal{L}(\theta) &= \mathbb{E}_{i \sim \mathcal{D}^{(p)}}\left[-p_i^\ast\log\hat{p}_i-(1-p_i^\ast)\log(1-\hat{p}_i)\right].
\end{aligned}
\end{equation}
Since GUI representations are often lengthy, we represent the state at step $t$ using the complete action history up to $t-1$ combined with the most recent screen observation:
\begin{equation}
\hat{s}_t = [a_1, a_2, \cdots, a_{t-1}, o_t].
\end{equation}
This form enables the progress model to effectively capture both task goal and sequential context necessary for accurate progress estimation.

\subsection{Online RL training}

We adapt the REINFORCE++ algorithm~\citep{hu2025reinforce++} to multi-turn
reinforcement learning. REINFORCE++  eliminates the need for a critic network and has demonstrated both efficiency and stability in training single-turn reasoning 
LLMs~\citep{xie2025logic, song2025r1}, making it well-suited for online RL training of LLM-based agents. To adapt to multi-turn
training of GUI agents, We follow the token-level credit assignment approach for multi-turn language agents proposed by 
\citet{wen2024reinforcing} to assign different reward discounts for inter-turn and intra-turn transitions.

\section{Experiments}
\label{sec:exp}

\subsection{Experimental settings}
\label{sub:exp_setting}

\paragraph{Environment}
We select the WikiHow benchmark~\citep{MobileEnv} to evaluate the effectiveness of \progrm{}. WikiHow is one of the few GUI interaction benchmarks that provide intermediate milestone rewards, which enables us to validate our proposed LCS-based key step discovery algorithm by comparing it against environment-reward-based key step discovery. The benchmark offers a canonical set of 577 annotated tasks for real-world GUI interactions within the WikiHow app. Of these, 150 tasks are used as the test set, while the remaining 427 tasks constitute the training set. Furthermore, according to \citet{MobileEnv}, the test set tasks are devided
into three categories: 
\begin{enumerate*}[label=(\arabic*)]
    \item Cross-Page tasks, where the agent needs to follow instructions to complete a series of navigations among 
        different pages,
    \item In-Page tasks, where the agent needs to find a specific article and perform some in-page operations like
        bookmarking, sharing, rating, \textit{etc}.\ according to the instructions,
    \item and QA tasks, where the agent needs to find a specific article and answer some questions according to it.
\end{enumerate*}
We follow this categorization and report results separately for the three types of tasks as well. 
\todo[disable]{describe three types of tasks.}

\paragraph{Reward model training data}
We used Qwen2.5-7B~\citep{Qwen2.5} and GPT-4o-mini\footnote{\url{https://platform.openai.com/docs/models/gpt-4o-mini}} to perform rollouts and collected 7,725 trajectories. To further augment the dataset and achieve a balanced distribution of steps between successful and failed trajectories, we applied data synthesis techniques, as detailed in \cref{sub:dat_synth}. This resulted in a final dataset of 10,438 trajectories, comprising 5,729 successful and 4,709 failed cases. For ORM training, we use the overall success or failure label of each trajectory. For \progrm{}, progress label for each step is generated using the pipeline described in \cref{sec:process_label}. Additional details regarding the reward model (RM) training data are provided in \cref{sec:RM_data_detail}.

\paragraph{Implementations}
We use Qwen2.5-7B as the base model for both reward models (RMs) and actors. Experiments are conducted with two variants of \progrm{s}, trained using either environment-reward-based progress labels or LCS-based progress labels, denoted as \ProgRM{\textsubscript{Env}} and \ProgRM{\textsubscript{LCS}}, respectively. Prior to reinforcement learning, the actor is initialized via supervised fine-tuning (SFT) for 10,000 steps using samples from the collected successful trajectories to acquire basic interaction abilities. We adapt the REINFORCE++ algorithm~\citep{hu2025reinforce++} to multi-turn reinforcement
learning of LLM-based agents following \citet{wen2024reinforcing} to add token-level credit assignment. We employ a remote environment server to enable parallel deployment of WikiHow alongside the RL trainer. The progress reward history length $k$ is set to 1 in the main experiments.

\paragraph{Baselines}
We use the trivial Outcome Reward Model (ORM) as the main baseline for comparison. \ProgRM{} is evaluated against ORM by assessing the performance of RL-finetuned actors (see \cref{sub:main_res}) as well as several direct metrics (see \cref{sub:anal_ablt}). We also include GUI-R1~\citep{GUIR1}, a step-level RL method that uses action-level matching with the ground truth as its reward signal, eliminating the need for reward models or hand-crafted GUI evaluation functions. For GUI-R1 training, 10,000 steps are sampled from the collected successful trajectories.\todo[disable]{Maybe GUI-R1 should be deleted.} In addition, we compare \ProgRM{-trained} agents with a series of recent state-of-the-art proprietary models, including Claude-3.7-Sonnet\footnote{\url{https://www.anthropic.com/claude/sonnet}}, GPT-4.1-mini\footnote{\url{https://platform.openai.com/docs/models/gpt-4.1-mini}}, and GPT-4o-mini\footnote{\url{https://platform.openai.com/docs/models/gpt-4o-mini}}.

\subsection{Results}
\label{sub:main_res}

\begin{table}[t]
\centering
\caption{Actor results on WikiHow task set. \ProgRM{\textsubscript{Env}} denotes \progrm{} trained with
    environment-reward-based progress labels and \ProgRM{\textsubscript{LCS}} denotes \progrm{} trained
    LCS-based progress labels. The Average Cumulative Rewards (Rwd) and Success Rates (SR, \%) are displayed. Both global 
    average results and per-category results are listed.}
\label{tab:main_rlt}
\begin{tabular}{ccccccccc}
    \toprule[1.5pt]
    \multirow{2.5}{*}{\textbf{Actor}} & \multirow{2.5}{*}{\textbf{Rwd}} & \multirow{2.5}{*}{\textbf{SR}} & \multicolumn{2}{c}{\textbf{Cross-Page}} & \multicolumn{2}{c}{\textbf{In-Page}} & \multicolumn{2}{c}{\textbf{QA}} \\
    \cmidrule(lr){4-5}\cmidrule(lr){6-7}\cmidrule(lr){8-9}
                                    &      &                   & \textbf{Rwd}  & \textbf{SR}                & \textbf{Rwd}  & \textbf{SR}                & \textbf{Rwd}  & \textbf{SR}                \\  
    \midrule
    GPT-4o-mini                     & 1.60 & 38.00             & 1.58 & 52.54             & 1.63 & 29.41             & 1.59 & \underline{27.50} \\ 
    GPT-4.1-mini                    & 2.16 & 52.00             & 2.13 & 71.19             & 2.47 & 47.06             & 1.81 & \textbf{30.00}    \\ 
    Claude-3.7-Sonnet               & 2.38 & 56.00             & 2.27 & \textbf{77.97}    & 2.78 & 58.82             & 2.03 & 20.00             \\ 
    \midrule
    Qwen2.5-7B                      & 1.89 & 31.33             & 1.71 & 54.23             & 2.08 & 15.69             & 1.91 & 17.50             \\  
    SFT                             & 2.32 & 56.00             & 1.95 & 62.71             & 3.02 & 84.31 & 1.98 & 10.00             \\  
    GUI-R1                          & 2.33 & 58.00             & 1.93 & 62.71             & 3.04 & \underline{86.27} & 2.02 & 15.00             \\
    w/ ORM                          & 2.35 & 58.67             & 2.05 & 72.88             & 2.96 & 78.43             & 2.00 & 12.50             \\
    w/ \ProgRM{\textsubscript{LCS}} & 2.37 & \underline{59.33} & 2.14 & 69.49             & 2.94 & 82.35             & 1.99 & 15.00             \\  
    w/ \ProgRM{\textsubscript{Env}} & 2.39 & \textbf{62.00}    & 2.12 & \underline{72.88} & 3.04 & \textbf{88.24}    & 1.95 & 12.50             \\ 
    \bottomrule[1.5pt] 
\end{tabular}
\end{table}

\todo[inline,disable]{1. Compare with SFT. How are models defeated by SFT on in-page tasks? 2. Compare with ORM. How can \progrm{s} defeat ORM on in-page tasks?}

The main results are presented in \cref{tab:main_rlt}. Actors trained with \progrm{} achieves the highest average cumulative rewards and success rates, with \ProgRM{\textsubscript{Env}} reaching 62.00\%. This performance surpasses all the state-of-the-art proprietary models using prompting (e.g., Claude-3.7-Sonnet at 56.00\%) as well as SFT actor trained with demonstrations (56.00\%). These baseline models often struggle to generalize to unseen environments and tend to become stuck, repeatedly outputting useless actions such as scrolling down. 

\begin{table}[t]
     \centering
     \caption{Accuracy of RM evaluations. The numbers are percentages (\%).
     Naive ORM holds an evidently higher false positive rate.}
     \label{tab:rm_eval_acc}
     \begin{tabular}{cccccccc}
         \toprule[1.5pt]
         \textbf{RM}                        & \textbf{\#TP}~$\lhookuparrow$  & \textbf{\#FN}~$\rhookdownarrow$  & \textbf{\#TN}~$\lhookuparrow$   & \textbf{\#FP}~$\rhookdownarrow$  & \textbf{Prec}~$\lhookuparrow$  & \textbf{Rec}~$\lhookuparrow$   & \textbf{Acc}~$\lhookuparrow$   \\
         \midrule
         ORM                          & 54.00 & 4.67 & 19.33 & 22.00 & 71.05 & 92.04 & 73.33 \\
         \ProgRM{\textsubscript{LCS}} & 57.33 & 2.00 & 33.33 & 7.33  & 88.66 & \textbf{96.63} & 90.67 \\
         \ProgRM{\textsubscript{Env}} & 57.33 & 4.67 & 36.67 & 1.33  & \textbf{97.73} & 92.47 & \textbf{94.00} \\
         \bottomrule[1.5pt]
     \end{tabular}
\end{table}

\ProgRM{} also outperforms ORM, especially on In-Page tasks. To further
understand the advantages of \progrm{} over ORM, we compare the direct
performance of the Reward Models (RM) in \cref{tab:rm_eval_acc} by matching
their predictions of success with the ground truth judgements from the
environment. The results show that \progrm{} consistently achieves stronger
correlation with groundtruth across all the metrics. In contrast, ORM exhibits
a significantly higher false positive rate, which undermines its reliability in
distinguishing between successful and unsuccessful trajectories.

\subsection{Analysis and ablation study}
\label{sub:anal_ablt}

\begin{table}[t]
    \centering
    \caption{Comparison of reward models (ORM, ORM\textsubscript{Claude} and \progrm{}) in terms of key step progress prediction error, average predicted final step score, and model inference latency.}
    \label{tab:rm_comp}
    \begin{tabular}{cccc}
        \toprule[1.5pt]
        \textbf{RM}                           & \textbf{Key Step Prog.\ Err}.\ $\rhookdownarrow$ & \textbf{Avg. Fin. Score} & \textbf{Latency (s)} $\rhookdownarrow$ \\
        \midrule
        ORM\textsubscript{Claude}                  & 0.638                 & 0.000           & 5.725       \\
        ORM\textsubscript{Claude}-CoT                  & 0.177                 & 0.593           & 8.531       \\
        ORM                          & 0.171                 & 0.595           & 0.050       \\
        \ProgRM{\textsubscript{LCS}} & 0.126                 & 0.846           & 0.050       \\
        \ProgRM{\textsubscript{Env}} & 0.036                 & 0.842           & 0.050       \\
        \bottomrule[1.5pt]
    \end{tabular}
\end{table} 


\todo[inline,disable]{Table comparing prediction error, final predicted rewards, and revocation latency.}

\todo[inline,disable]{Explain why QA performance can be improved with a larger estimation error. Because the base performance is too low.}

\paragraph{Analysis of per-category performances}
We observe that, except for the model trained with \ProgRM{\textsubscript{Env}}, RL-finetuned models do not achieve higher scores on In-Page tasks compared to the SFT baseline. By looking through the actors' trajectories, 
it is found that the new failures of In-Page tasks after being finetuned with ORM and \ProgRM{\textsubscript{LCS}} lie in 
misactions, \textit{e.g.}, the instruction requires giving a thumb-up to an article, but the actor gives a thumb-down.
Such misactions can be attributed to the misleading of RMs to some extent, as it is further noticed that 
ORM and \ProgRM{\textsubscript{LCS}} may assign such results a high score in some cases. This is
consistent with the higher false positive rates of \ProgRM{\textsubscript{LCS}} and ORM in \cref{tab:rm_eval_acc}.


For QA tasks, closed-source proprietary models achieve the best performances. Interestingly, the performance of finetuned models on QA tasks is lower than that of the base model prior to SFT. Completing a QA task requires the agent to generate an appropriate answer based on a reference article; thus, proprietary models with superior natural language capabilities excel in this category. In contrast, SFT focused on GUI-specific tasks somewhat diminishes the actor’s general language ability, resulting in a lower baseline for QA performance. Subsequent RL finetuning partially improves this, but does not fully restore the original performance level. We anticipate that additional RL fine-tuning steps may further enhance the actor's capabilities on QA tasks.

We additionally compare \progrm{} with ORM and a general-purpose evaluator~\cite{DigitalAgentEvaluator} based on Claude-3.7-Sonnet, referred to as ORM\textsubscript{Claude}, for progress estimation. The comparison includes progress estimation error for key steps, the average predicted score for final steps, and invocation latency. Results are presented in \cref{tab:rm_comp}. We evaluate both ORM\textsubscript{Claude}-s with and without Chain-of-Thought (CoT) reasoning; the used prompts are provided in \cref{sec:prompt}.

\paragraph{Key step progress estimation} We collect actor trajectories on WikiHow and consider the steps
receiving environment milestone rewards to be ground-truth key steps. The progress labels are then assigned
to the key steps by assuming the progress gains are even among them. Then we leverage various types of RMs 
to estimate the progress of these key steps, and calculate the mean absolute error. The results are presented in the second column of \cref{tab:rm_comp}. \ProgRM{} achieves the lowest estimation errors among all the reward models evaluated, indicating its ability to produce more accurate progress estimations and provide finer-grained guidance during actor training. Notably, the estimation error of \ProgRM{\textsubscript{LCS}} is significantly higher than that of \ProgRM{\textsubscript{Env}}, suggesting that there are still gaps between LCS-based and environment-reward-based key step discovery. Further optimization of automatic key step discovery algorithms remains an important direction for future work. Naively trained ORM acquires a comparable estimation error with an out-of-box general-purpose evaluator with CoT, not
showing an evident advantage. This phenomenon proves that trivial ORM training cannot endow the RM with the capability of
GUI task progress estimation.

\paragraph{Final step score prediction}
Different types of reward models are used to predict scores for the final steps of trajectories, as shown in \cref{tab:rm_comp}. \ProgRM{\textsubscript{Env}} and \ProgRM{\textsubscript{LCS}} yield similar average final scores. Likewise, the average final score predictions for ORM and ORM\textsubscript{Claude}-CoT are also comparable. The average score predicted by ORM is noticeably lower than that of \progrm{}. This is because ORM tends to assign scores close to either 0 or 1, effectively functioning as a binary indicator of trajectory success or failure. In contrast, \progrm{} estimates the agent’s cumulative task progress, so the scores for failed trajectories are not necessarily close to zero. It is important to note that even within a failed trajectory, there may be positive steps that contribute toward the task goal. In such cases, the coarse 0–1 scoring provided by ORM fails to appropriately reward these intermediate achievements, inadequately penalizing or ignoring the agent’s partial progress. In contrast, \progrm{} can assign moderate credit to these steps, encouraging more effective and efficient exploration during RL training. Additionally, we observe that ORM\textsubscript{Claude} is unable to generate meaningful scores and is therefore unsuitable for agent evaluation or training.

\paragraph{Invocation Latency}
The invocation latencies for different types of reward models are shown in \cref{tab:rm_comp}. The lightweight, self-hosted RMs are efficient and well-suited for online training, with response times suitable for practical use. In contrast, invoking ORM\textsubscript{Claude} incurs significantly higher latency. Even without Chain-of-Thought (CoT) reasoning, ORM\textsubscript{Claude} requires several seconds to return a response, and the latency increases further with CoT enabled. This makes accessing general-purpose evaluators such as ORM\textsubscript{Claude} entirely impractical for online training.

\begin{figure}[t]
    \centering
    \begin{minipage}[]{0.54\linewidth}

        \centering
        
    \hspace{-0.9cm}
        \includegraphics[width=\linewidth]{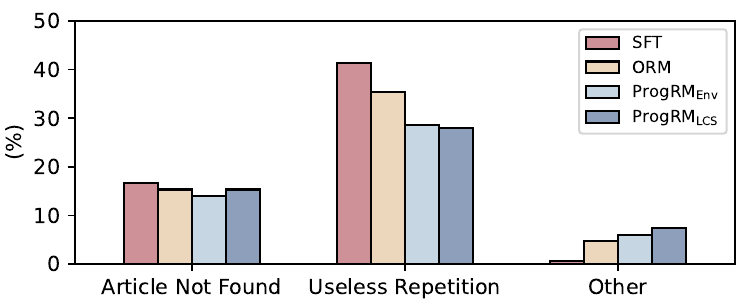}
        \caption{Actor failure mode analysis}
        \label{fig:fail_mod}
    \end{minipage}
    ~
    \begin{minipage}[]{0.44\linewidth}
        \centering
        \captionof{table}{Ablation study about history length $k$ of progress reward and 
        training steps.
        }
        \label{tab:hist_len_ablt}
        \begin{tabular}{cccc}
            \toprule[1.5pt]
            $k$ & \textbf{Train.\ Step}.\ & \textbf{Rwd} & \textbf{SR} (\%) \\
            \midrule
            $k=1$ & $\sim$20K & 2.39 & 62.00 \\
            $k=3$ & $\sim$20K & 2.30 & 54.00 \\
            $k=1$ & $\sim$30K & 2.43 & 67.33 \\
            \bottomrule[1.5pt]
        \end{tabular}
    \end{minipage}
\end{figure}

\paragraph{Actor failure mode analysis}
\todo[inline,disable]{Error type analysis}
We summarize two typical modes of failure of the actors, \textit{i.e.}, ``article not found'' and
``useless repetition''. ``Article not found'' is referred to as the error where the agent fails to figure out the proper search keywords to reach a target article page in WikiHow app. ``Useless repetition'' indicates that the agent
repeats some useless actions without achieving any actual progress to complete the task.
Statistics are performed on these error modes and depicted in \cref{fig:fail_mod}.
Compared to SFT and ORM-trained actors, the ``useless repetition'' failures of \progrm{}-trained actors 
decrease most remarkably. By training with \progrm{}, the actor learns to perform actions that can result
in most progress gains and prevent useless repetitive actions that bring no new progress. 

\todo[inline,disable]{Place figure of mode analysis and small table of ablation in one line.}

\paragraph{Ablation study about history length $k$ and training steps}
\todo[inline,disable]{Ablation about history length $k$.}
We conduct ablation study about the history length $k$ of progress reward (see \cref{sub:prog_rwd_def}) 
used in RL training. As shown in \cref{tab:hist_len_ablt}, increasing history length degrades the actor's
performance significantly, revealing that increasing history length is not suitable for the specific GUI interaction
tasks. Progress reward with history length $k>1$ gives a step the credit for the cumulative progress gain of
$k$ contiguous actions. This may be useful for some cases where per-step progress gain is little while a group
of contiguous actions can result in a relatively meaningful progress gain. Such cases usually mean particularly
long episodes or overly atomic action space, which is not case for common GUI interaction environments.
We further conduct a supplementary experiment by training the GUI agent with \ProgRM{\textsubscript{Env}} for more steps 
and demonstrate the result in \cref{tab:hist_len_ablt}. A longer training process further boosts the actor's performance, increasing the success rate from 62.00\% to \textbf{67.33\%}. This demonstrates the potential of \progrm{} for continuously enhancing actor performance.





\section{Related works}

\paragraph{Auto-evaluation for GUI agents} Strong capability of LLMs reveals the feasibility of enabling
auto-evaluation of GUI interaction instead of hand-crafted evaluation
functions.  \citet{DigitalAgentEvaluator} systematically summarizes the
framework of LLM\slash VLM-based auto-evaluators for GUI agents. This framework
is popularly used in a series of
instruction-first~\citep{PAE,LearnByInteract,Explorer} and
trajectory-first~\citep{NNetnav} data augmentation works. However, the reliance
on expensive and high-latency super models makes it stressful to be afforded
and leveraged in online training. In contrast, \citet{WebRL} adopts a more
lightweight ORM in RL (Reinforcement Learning) training.  Nevertheless, ORM
training fails to exploit the intermediate steps in training trajectories and
cannot predict accurate progress during interaction. Therefore, we present
\progrm{} to sufficiently exploit all the training steps and provide the actor
with meticulous guidance by predicting episode progress.

\paragraph{RL for LLM-based GUI agents} GUI interaction is a typical decision-making
problem and RL methods have been explored by the community. \citet{DigiRL,DigiQ} mainly
exploit static trajectory datasets to conduct offline learning. \citet{VEM} leverages
a general-purpose evaluator to provide reward and train a Value Environment Model
to avoid direct accessing an online GUI environment. WebRL~\citep{WebRL} and
DistRL~\citep{DistRL} explore online training for GUI interaction tasks. Outcome reward models
are used to produce rewards during online training. Except for normal trajectory-level
RL, recent works~\citet{UIR1,GUIR1} also explore step-level RL for GUI tasks, inspired by the success of 
DeepSeek-R1~\citet{DeepSeekR1} on reasoning tasks. In this paper, we propose a new
process reward model for GUI interaction tasks, \progrm{}, to provide exquisite
progress reward in RL training. Ideally, \progrm{} can be combined with any trajectory-level
RL methods.

\paragraph{ORMs and PRMs in reasoning tasks} Outcome Reward Model~(ORM) and
Process Reward Model~(PRM) has been widely used in reasoning tasks like
mathematical problems, coding tasks, \textit{etc}.\ for verification-guided
generation~\citep{GSM8K,YifeiLi2023ACL_StepAwareVerifier,FeiYu2023_MathORM,VerifyStepByStep},
reinforcement learning~\citep{MathShepherd,MRT}, and preference
learning~\citep{ImplicitPRM}. ORM is generally trained according to the final
answers of reasoning problems, which are commonly easy to obtain.
\citet{VerifyStepByStep} trains PRM using human-annotated process labels, which
are overly costly. \citet{MathShepherd} proposes to use Monte-Carlo search to
estimate the likelihood of reaching the correct answer starting from an
intermediate state. However, PRMs trained with such labels are more like a
cumulative expected reward function (value function) rather than a reward function
grading the instant state. \citet{ImplicitPRM} proposes to derive a PRM from an
ORM for preference learning. Similarly, the probability gain of reaching the
correct answer during transition between two states is measured by an ORM and
used as process reward in \cite{MRT}. Unlike single-turn answer generation for
reasoning tasks, GUI interaction is always in multiple turns. As a PRM for GUI
agents, \progrm{} measures the value of a complete interaction step rather than an
incomplete state during a single generation. A new algorithm is also developed
to discover {\em key steps} in trajectories and assign progress labels
accordingly for \progrm{}.

\paragraph{Progress reward} \citet{ELE} proposes {\em progress reward} to
measure the effectiveness of agents' actions and guide agents' exploration
during RL training. The progress reward function is trained using hundreds of
millions of human-playing steps on NetHack~\citep{NetHackDataset}. In contrast,
\progrm{} is dedicated to GUI interaction and trained with agent-explored
trajectories, reducing the workload of human annotation. \citet{MRT} derives
progress reward from the perspective of minimizing cumulative regret and uses
it to improve the efficiency of math problem solving. It leverages an ORM to
measure the probability of reaching the correct answer from a partial solution.
In comparison, \progrm{} is designed to apply to a complete GUI interaction
step and is trained with meticulous progress labels to attain more accurate
progress estimation rather than directly borrowing an ORM.

\section{Conclusion}

In this work, we introduce \progrm{}, a novel reward model for GUI agents that provides fine-grained reward signals during online RL training by accurately estimating progress at each step. We also propose an efficient self-annotation algorithm to generate appropriate progress labels for \progrm{} training. Agents trained with \progrm{} outperform both proprietary LLM-based agents and those trained with conventional ORMs, demonstrating the strong effectiveness and potential of our approach.

\bibliographystyle{plainnat}
\bibliography{neurips_2025}

\appendix


\section{Details of soft \& hard LCS algorithms}
\label{sub:slcs_def}

\todo[inline,disable]{motivation of soft LCS: better handle actions with natural languages; hard LCS often results in overly short result
sequences.}
The soft Longest Common Subsequence (LCS) algorithm is proposed to better handle actions with natural 
language arguments, which are unsuitable 
for direct exact match. It is derived from the standard ``hard'' LCS algorithm by replacing
the exact equation with a soft match function. To be specific, given two sequences $\mathbf{a} = \{a_i\}_{i=1}^m$
and $\mathbf{b} = \{b_j\}_{j=1}^n$, 
the LCS of $\mathbf{a}$ and $\mathbf{b}$, $\mathtt{LCS}(\mathbf{a}, \mathbf{b})$,
and its length can be solved by dynamic programming. The Bellman equation is as follows.
\begin{equation}
    |\mathtt{LCS}(\mathbf{a}_{1:i}, \mathbf{b}_{1:j})| = \begin{cases}
        |\mathtt{LCS}(\mathbf{a}_{1:i-1}, \mathbf{b}_{1:j-1})|+1 & a_i = b_j \\
        \max\{|\mathtt{LCS}(\mathbf{a}_{1:i-1}, \mathbf{b}_{1:j})|, |\mathtt{LCS}(\mathbf{a}_{1:i}, \mathbf{b}_{1:j-1})|\} & a_i\ne b_j. \\
    \end{cases}
\end{equation}
By replacing the hard match (\textit{i.e.}, exact equation) with a soft match function $f$, we obtain the
Bellman equation for soft LCS algorithm:
\begin{equation}
    \begin{aligned}
        \mathtt{SoftLCS}(\mathbf{a}_{1:i}, \mathbf{b}_{1:j}) = \max\{ &\mathtt{SoftLCS}(\mathbf{a}_{1:i-1}, \mathbf{b}_{1:j-1})+f(a_i, b_j), \\
        &\mathtt{SoftLCS}(\mathbf{a}_{1:i-1}, \mathbf{b}_{1:j}), \mathtt{SoftLCS}(\mathbf{a}_{1:i}, \mathbf{b}_{1:j-1})\}. \\
    \end{aligned}
\end{equation}
The soft match function $f$ is defined according to the particular action space. Given two WikiHow actions,
$a = (\mathtt{type}_a, \mathtt{element}_a, \mathtt{text}_a)$, $b = (\mathtt{type}_b, \mathtt{element}_b, \mathtt{text}_b)$,
$f$ is defined as
\begin{equation}
    f(a, b) = \begin{cases}
        0 & \mathtt{type}_a \ne \mathtt{type}_b \\
        \mathtt{SBERT}(\mathtt{text}_a, \mathtt{text}_b) & \mathtt{type}_a=\mathtt{type}_b \in \{\mathtt{INPUT}, \mathtt{ANSWER}\} \\
        \varepsilon & \mathtt{type}_a=\mathtt{type}_b=\mathtt{NOTHING} \\
        1[a=b] & \text{otherwise}, \\
    \end{cases}
\end{equation}
\todo[disable]{Add some explanation. NOTHING is not completely disabled as it may be a necessary waiting. replace 0.4 with a variable.}
where $\mathtt{SBERT}$ denotes computing text similarity with Sentence Transformer~\citep{SentenceTransformer} 
and 0.4 is used for $\varepsilon$.
This soft match function gives a soft weight for actions with free-form natural language arguments can lead to
more finegrained similarity. Besides, the function penalizes the match of empty actions \verb|NOTHING| so
that more weights are assigned to the other actual actions and a more meaningful match can be obtained. Note that
the match of \verb|NOTHING| is not completely disabled as some of them may be necessary waiting that should 
be preserved.

\todo[inline,disable]{describe how we compute group LCSs.}
Another problem is how to compute LCS for multiple sequences with a group. As a direct application of
dynamic programming to LCS computation of more than two sequences is too complex, we adopt the two-sequence LCS
algorithm to achieve approximation, \textit{i.e.},
\begin{equation}
    \widetilde{\mathtt{LCS}}(\mathbf{a}_1, \mathbf{a}_2, \cdots, \mathbf{a}_n) = \mathbf{a}_1 \odot \mathbf{a}_2 \odot \cdots \odot \mathbf{a}_n.
\end{equation}
Here we adopt the left-associative binary operator $\odot$ to denote two-sequence LCS function for the convenience of 
expression.

The similarity threshold for trajectory grouping $\theta_L$ is 0.6 in our experiments.

\section{RM training data synthesis}
\label{sub:dat_synth}

\begin{figure}
    \centering
    \subfigure[Trajectory number statistics of collected RM training data on WikiHow \label{fig:wkh_traj_stat}]{\includegraphics[width=\linewidth]{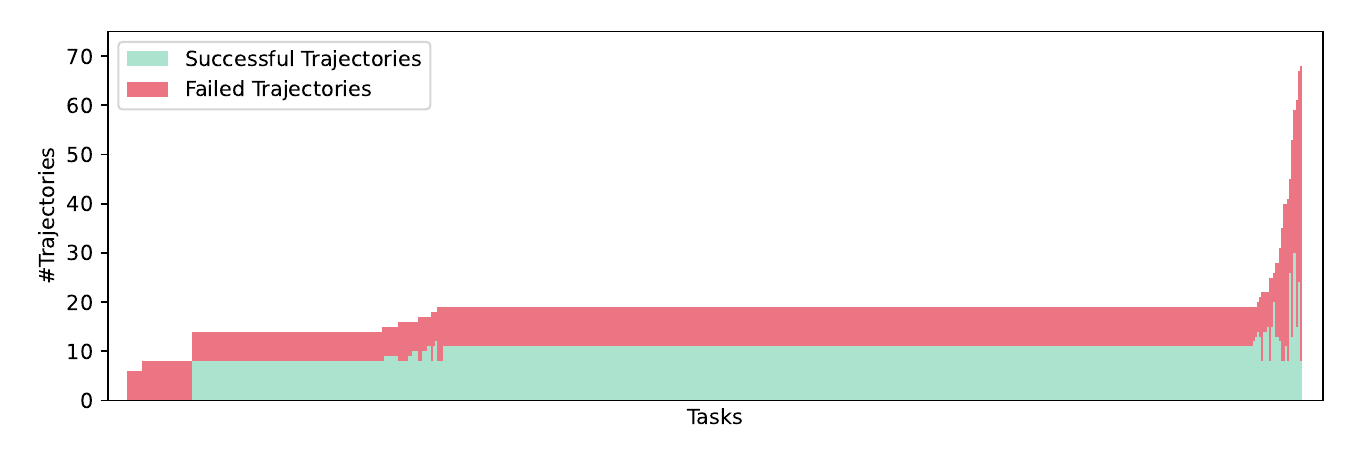}}

    \subfigure[Step number statistics of collected RM training data on WikiHow \label{fig:wkh_step_stat}]{\includegraphics[width=\linewidth]{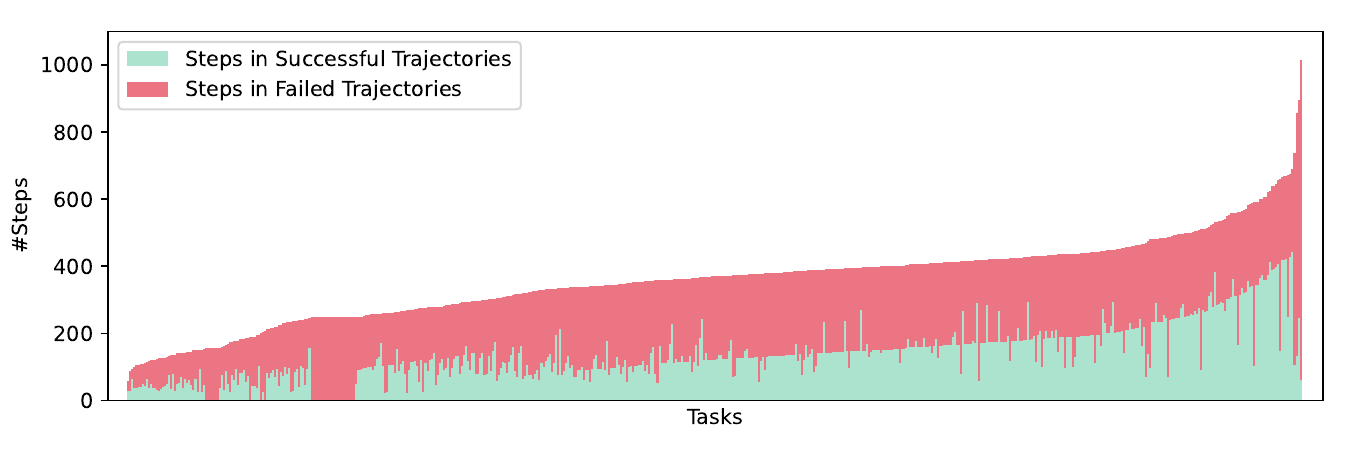}}
    \caption{Statistics of the collected reward model (RM) training data for WikiHow. \Cref{fig:wkh_traj_stat} displays the number of successful and failed trajectories, with a success-to-failure ratio of approximately 1.22. \Cref{fig:wkh_step_stat} presents the number of steps in successful versus failed trajectories, with a step ratio of about 0.63.}
    \label{fig:dat_blnc}
\end{figure}

To supplement the collected trajectories and achieve a better data balance, we perform data synthesis based on collected agent trajectories.

\paragraph{Failed trajectory synthesis} Failed trajectories are synthesized in two ways, 
\begin{enumerate*}[label=\arabic*)]
    \item combining mismatched instruction and action trajectory, \textit{e.g.}, combining instruction for task
    $a$ with execution trajectory of task $b$
    \item and leveraging a random walk trajectory.
\end{enumerate*}

\paragraph{Successful trajectory synthesis} Successful trajectories are synthesized based on ``prototype'' 
trajectories, \textit{i.e.}, given an existing successful trajectory for a particular task, a new successful trajectory can be generated by removing or adding empty or effectless action tuples. For example, in the WikiHow environment, an empty action corresponds to the action \verb|NOTHING|, while effectless action tuples might include actions such as scrolling down followed by scrolling up, or going back and immediately repeating the last action. If the agent's prior exploration fails to produce any successful trajectories for a given task, a successful trajectory is manually annotated by the authors.

Using this synthesis approach, the final dataset consists of 5,729 successful trajectories and 4,709 failed trajectories. The dataset statistics are presented in \cref{fig:dat_blnc}.

\section{Details of training data for reward models}
\label{sec:RM_data_detail}

We leveraged Qwen2.5-7B and GPT-4o-mini to auto-collect a total of 10,438 trajectories, consisting of 5,729 successful and 4,709 failed trajectories. These trajectories comprise 207,102 steps in total, with 79,718 steps originating from successful trajectories and 127,384 from failed ones.
The trajectories are partitioned into subsets based on their task goals, resulting in a training set of 7,175 trajectories, a validation set of 1,751 trajectories, and a test set of 1,512 trajectories. This trajectory data can be used directly for naive ORM training.
For \progrm{} training, we further split the trajectories into individual steps. All steps from successful trajectories are retained, while 62.58\% of steps from failed trajectories are sampled to balance the success-to-failure step ratio at approximately 1:1. This results in a \progrm{} training set of 113,270 steps, a validation set of 15,935 steps, and a test set of 30,220 steps.

\section{Experiment details}
\label{sec:exp_detail}

All experiments are conducted on a single machine equipped with 8 NVIDIA A800 GPUs of 80 GB memory. We use the Adam optimizer with $(\beta_1, \beta_2) = (0.9, 0.95)$ for both reward model training and online RL training of agent models. DeepSpeed ZeRO is employed to optimize GPU memory usage during training. The hyperparameters for training the reward models, including ORM, \ProgRM{\textsubscript{Env}}, and \ProgRM{\textsubscript{LCS}}, are listed in \Cref{tab:rm_hyperparams}. Hyperparameters for online RL agent training are provided in \Cref{tab:agent_rl_hyperparams}. Training a reward model takes roughly 3 hours, while agent RL training requires around 20 hours.

\begin{table}[h]
    \centering
    \caption{Hyperparameters for reward model training}
    \label{tab:rm_hyperparams}
    \begin{tabular}{l l}
        \toprule
        \textbf{Hyperparameter}   & \textbf{Value} \\
        \midrule
        Learning rate     & $5 \times 10^{-5}$ \\
        Batch size        & 128 \\
        Epoch             & 3 \\
        LR scheduler type & Cosine \\
        Warmup ratio      & 0.03 \\
        ZeRO stage        & 2 \\
        \bottomrule
    \end{tabular}
\end{table}

\begin{table}[h]
    \centering
    \caption{Hyperparameters for agent RL training}
    \label{tab:agent_rl_hyperparams}
    \begin{tabular}{l l}
        \toprule
        \textbf{Hyperparameter}   & \textbf{Value} \\
        \midrule
        Learning rate         & $7 \times 10^{-5}$ \\
        Batch size            & 64 \\
        Epoch                 & 1 \\
        LR scheduler type     & Cosine \\
        Warmup ratio          & 0.03 \\
        KL coefficient        & 0.01 \\
        Gamma                 & 0.9 \\
        PPO clip              & 0.2 \\
        Rollout temperature   & 1.0 \\
        ZeRO stage            & 3 \\
        \bottomrule
    \end{tabular}
\end{table}

\section{Details of WikiHow deployment}
\label{sub:mobenv_deploy}

We implement a RESTful remote environment server to enable parallel deployment of WikiHow along with the RL trainer.
For convenience, we use two Docker images to host the Android Emulator\texttrademark{} and the replay server
for WikiHow environment~\citep{MobileEnv}. Flask\footnote{\url{https://flask.palletsprojects.com/en/stable/}} is 
used to build the main server of a remote environment. To reduce the communication latency, only the prompts are 
transferred on an HTTP (Hyper Text Transport Protocol) flow. To control consumed computing resources and achieve
as efficient a simulation as possible, a daemon management thread is implemented to 
create emulator instances, monitor their running state, and clean stale instances in time.
The remote environment server is deployed on a CentOS machine with KVM (Kernel-Based Virtual Machine) enabled,
CPU of 64 virtual threads, 1.97 TiB of memory, and an A800 GPU equipped.

\ifwithsupp
	\ifwithsupp
	\newcommand{\Sthirtythree}{\cref{sub:anal_ablt}}
\else
	\newcommand{\Sthirtythree}{\S~3.3 in the main paper}
\fi

\section{Case study}

\begin{figure}[h]
	\centering
	\includegraphics[width=\linewidth]{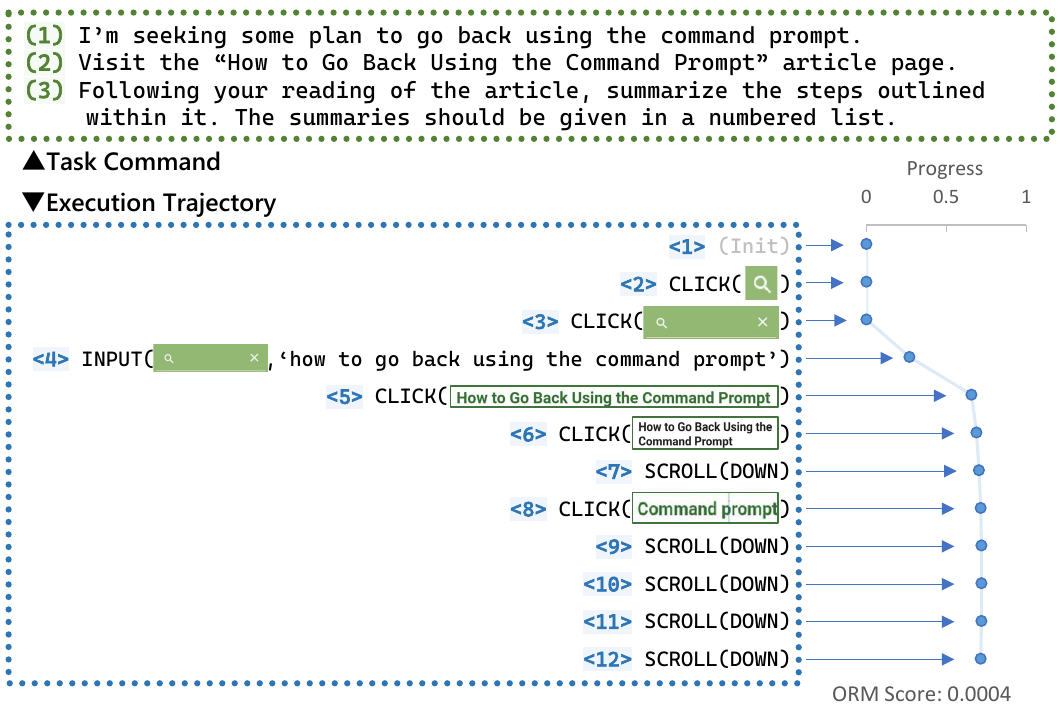}
	\caption{A failed trajectory with partial progress showing with the
		progress scores predicted by \progrm{} and the final score predicted by
		ORM.  Each line in the bottom-left section is an action taken by the
		agent in episode. The progress values predicted by \progrm{} after each
		action executed are illustrated in the bottom-right section. The agent
	achieves partial progress in the episode, while doesn't reach the final
goal and stops at a progress score lower than 1 (100\%).  \texttt{Init} in the
figure is not an actual action, but a placeholder.}
	\label{fig:orm_inadeq_penal}
\end{figure}

\todo[inline,disable]{Some callback to the over-penalization of ORM}

In this section, we give some cases to show the potential of \progrm{} for
predicting moderate progress scores and assigning adequate credits for
interaction steps.

\paragraph{Over-penalization of ORM} As stated in {\em Final step score
prediction} of \Sthirtythree{}, ORM predicting a single less informative score
at the episode end indicating mere success or failure can potentially
over-penalize the effective steps in a failed trajectory and leads to
insufficient exploitation of failed trajectories. As a comparison, \progrm{}
measures an adequate progress score for each step and can assign proper credits
for steps even in failed trajectories. As illustrated in
\cref{fig:orm_inadeq_penal}, the agent completed partial task without achieving
the final goal. The score predicted by the trivial ORM marks the whole
trajectory as failed inadequately penalizes all the steps in the trajectory,
although some steps do cause meaningful progress gains (actions \verb|<4>| and
\verb|<5>| in \cref{fig:orm_inadeq_penal}).  \todo[disable]{Keep action number
format the same as in the figure. And add some explanation about the action per
line.} In contrast, the progress curve predicted by \progrm{} accurately
reflects the effect of the actions. Thus, \progrm{} can assign moderate credits
to these valuable actions even in a failed trajectory.  \todo[disable]{Check
the index.}

\begin{figure}[t]
	\centering
	\subfigure[A successful trajectory and the temporal variation of
	\progrm{}-measured progress.
\label{fig:norm_succ_traj}]{\includegraphics[width=0.77\linewidth]{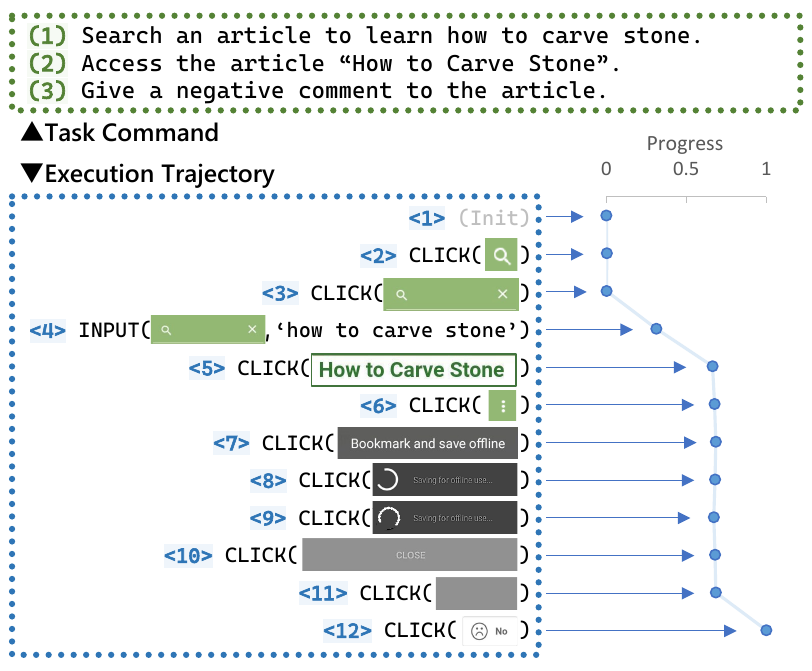}}

	\subfigure[A successful trajectory with agent's hesitation (\texttt{GOBACK}
	at step \texttt{<5>}) and the temporal variation of \progrm{}-measured
progress.
\label{fig:hesit_succ_traj}]{\includegraphics[width=0.77\linewidth]{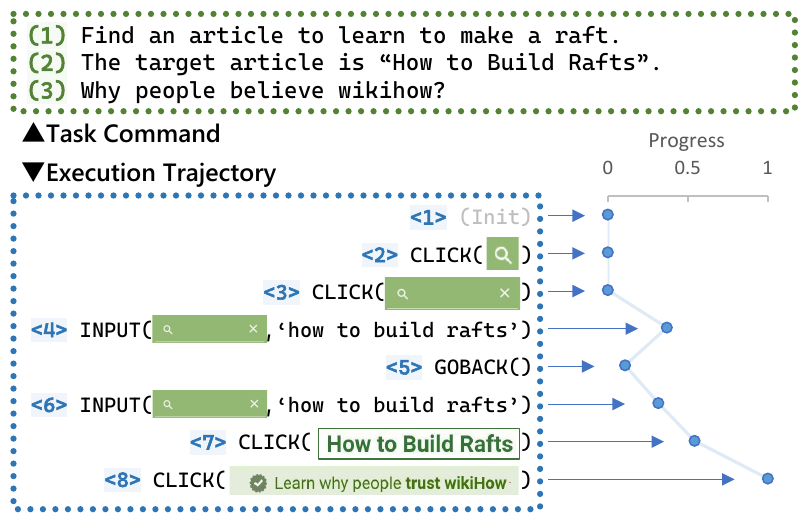}}
	\caption{Temporal variation of progress measurement over successful
	episodes. Each line in the bottom-left sections is an action taken by the
agent in episode. The progress scores predicted by \progrm{} after each action
executed are illustrated in the bottom-right sections.}
	\label{fig:succ_trajs}
\end{figure}

\paragraph{Temporal variation of progress measurement over successful episodes}
We further show the capacity of \progrm{} for task progress estimation with two
successful trajectories. \Cref{fig:norm_succ_traj} demonstrates a successful
trajectory with its progress-step curve. The agent completes the task
progressively, accompanied by that the progress score increases progressively
to 1 (100\%). The key steps achieving sub-goals (steps \verb|<4>|, \verb|<5>|,
and \verb|<12>| in \cref{fig:norm_succ_traj}) and non-key steps are clearly
distinguished through the corresponding progress gains, with higher gains
corresponding to key steps and lower gains corresponding to non-key steps,
\todo[disable]{Check the indices} revealing the capacity of \progrm{} for
identifying the valuable actions.  \Cref{fig:hesit_succ_traj} illustrates a
successful trajectory where the agent hesitates with a \verb|GOBACK| action
(action \verb|<5>| in \cref{fig:hesit_succ_traj}) \todo[disable]{Check the
indices} after search but recovers later.  \ProgRM{} accurately catches the
progress fluctuation and reflects it in the curve by a progress decline and the
following rebound. In such way, \progrm{} manages to assign a proper credit for
each steps in the trajectory and thus provide more exquisite guidance during
actor training.

\fi

\section{Prompts used in experiments}
\label{sec:prompt}

The used prompts for RMs and actors are listed in \cref{tab:progrm_prompt}, \cref{tab:orm_prompt}, \cref{tab:srm_prompt},
\cref{tab:srm_cot_prompt}, and \cref{tab:actor_prompt}.

\begin{longtable}{rp{11.5cm}}
	\caption{Prompt for \progrm{}}
	\label{tab:progrm_prompt} \\

	\toprule[1.5pt]
	\endhead

	\bottomrule[1.5pt]
	\endfoot

	\textcolor{systemcolor}{System: }
	& You are an expert of mobile use, especially the app of WikiHow. This is a public and popular wiki platform you surf everyday. You know well how people can search for specific articles, check author information, find more related articles, make bookmarks, etc. on WikiHow app. So you can accurately assess how efficient the people are finishing a particular task on this app. \\
 	& \\
	& Now you will be given a trajectory of other's operation, including \\
 	& \\
	& \textasteriskcentered{} instructions describing the task goal \\
	& \textasteriskcentered{} history actions leading to current state \\
	& \textasteriskcentered{} screen representation reflecting the current state \\
 	& \\
	& You should give a percentage which is an estimation of his progress on this task. \\
	\textcolor{usercolor}{User: }
	& Instructions: \\
	& \texttt{\$\{instructions\}} \\
	&  \\
	& History Actions: \\
	& \texttt{\$\{actions\}} \\
	&  \\
	& Current screen: \\
	& \texttt{\$\{screen\}} \\
\end{longtable}

\begin{longtable}{rp{11.5cm}}
	\caption{Prompt for ORM}
	\label{tab:orm_prompt} \\

	\toprule[1.5pt]
	\endhead

	\bottomrule[1.5pt]
	\endfoot

	\textcolor{systemcolor}{System: }
	& You are an expert of mobile use, especially the app of WikiHow. This is a public and popular wiki platform you surf everyday. You know well how people can search for specific articles, check author information, find more related articles, make bookmarks, etc. on WikiHow app. So you can accurately assess how efficient the people are finishing a particular task on this app. \\
 	& \\
	& Now you will be given a trajectory of other's operation, including \\
 	& \\
	& \textasteriskcentered{} instructions describing the task goal \\
	& \textasteriskcentered{} history actions leading to current state \\
	& \textasteriskcentered{} screen representation reflecting the current state \\
 	& \\
	& You should give a judgment about the success or failure of this task. \\
	\textcolor{usercolor}{User: }
	& Instructions: \\
	& \texttt{\$\{instructions\}} \\
	&  \\
	& History Actions: \\
	& \texttt{\$\{actions\}} \\
	&  \\
	& Current screen: \\
	& \texttt{\$\{screen\}} \\
\end{longtable}

\begin{longtable}{rp{11.5cm}}
	\caption{Prompt for ORM\textsubscript{Claude}}
	\label{tab:srm_prompt} \\

	\toprule[1.5pt]
	\endhead

	\bottomrule[1.5pt]
	\endfoot

	\textcolor{systemcolor}{System: }
	& You are an expert of mobile use, especially the app of WikiHow. This is a public and popular wiki platform you surf everyday. You know well how people can search for specific articles, check author information, find more related articles, make bookmarks, etc. on WikiHow app. So you can accurately assess how efficient the people are finishing a particular task on this app. \\
 	& \\
	& Now you will be given a trajectory of other's operation, including \\
 	& \\
	& \textasteriskcentered{} instructions describing the task goal \\
	& \textasteriskcentered{} history actions leading to current state \\
	& \textasteriskcentered{} screen representation reflecting the current state \\
 	& \\
	& You should give a percentage which is an estimation of his progress on this task. You should directly give your answer. Do not output any needless thoughts or explanations. \\
	\textcolor{usercolor}{User: }
	& Instructions: \\
	& \texttt{\$\{instructions\}} \\
	&  \\
	& History Actions: \\
	& \texttt{\$\{actions\}} \\
	&  \\
	& Current screen: \\
	& \texttt{\$\{screen\}} \\
\end{longtable}

\begin{longtable}{rp{11.5cm}}
	\caption{Prompt for ORM\textsubscript{Claude}-CoT}
	\label{tab:srm_cot_prompt} \\

	\toprule[1.5pt]
	\endhead

	\bottomrule[1.5pt]
	\endfoot

	\textcolor{systemcolor}{System: }
	& You are an expert of mobile use, especially the app of WikiHow. This is a public and popular wiki platform you surf everyday. You know well how people can search for specific articles, check author information, find more related articles, make bookmarks, etc. on WikiHow app. So you can accurately assess how efficient the people are finishing a particular task on this app. \\
 	& \\
	& Now you will be given a trajectory of other's operation, including \\
 	& \\
	& \textasteriskcentered{} instructions describing the task goal \\
	& \textasteriskcentered{} history actions leading to current state \\
	& \textasteriskcentered{} screen representation reflecting the current state \\
 	& \\
	& You should give a percentage which is an estimation of his progress on this task. You should first generate an explicit thought and then give your answer. You should output in the following format: \\
	&  \\
	& \texttt{<think>} \\
	& some thoughts \\
	& \texttt{</think>} \\
	& \texttt{<answer>} \\
	& 0.42 \\
	& \texttt{</answer>} \\
	&  \\
	& Follow the format above strictly. And note that the example above are just an example demonstrating the output format and takes NO ANY RELATION with the following inputs. \\
	\textcolor{usercolor}{User: }
	& Instructions: \\
	& \texttt{\$\{instructions\}} \\
	&  \\
	& History Actions: \\
	& \texttt{\$\{actions\}} \\
	&  \\
	& Current screen: \\
	& \texttt{\$\{screen\}} \\
\end{longtable}

\begin{longtable}{rp{11.5cm}}
	\caption{Prompt for actors}
	\label{tab:actor_prompt} \\

	\toprule[1.5pt]
	\endhead

	\bottomrule[1.5pt]
	\endfoot

	\textcolor{systemcolor}{System: }
	& You are a clever mobile assistant. You are very familiar with WikiHow and can navigate its app expertly. Now you will be given several information about the task and the screen at the current step, and you need to take an appropriate action according to the given information to finish the task in STEP steps. The action should in format of Python function call. Available actions are: \\
	&  \\
	& {\tt \textasteriskcentered{} INPUT(element\_id: int, text: str) \# You can input something into text box through this action} \\
	& {\tt \textasteriskcentered{} CLICK(element\_id: int) \# You can click on some clickable element through this action} \\
	& {\tt \textasteriskcentered{} LONG\_CLICK(element\_id: int) \# You can long lick on some clickable element through this action} \\
	& {\tt \textasteriskcentered{} SCROLL(direction: Enum) \# You can scroll UP/DOWN/LEFT/RIGHT to browse long/wide pages through this action} \\
	& {\tt \textasteriskcentered{} ANSWER(text: str) \# You can generate an answer to me through this action} \\
	& {\tt \textasteriskcentered{} GOBACK() \# You can go back to the previous screen by pressing GOBACK button of mobile} \\
	& {\tt \textasteriskcentered{} DO\_NOTHING() \# You can do nothing and just wait for a step} \\
	&  \\
	& Here are some examples of actions: \\
	&  \\
	& \texttt{{`}{`}{`}} \\
	& \texttt{INPUT(3, "scooter")} \\
	& \texttt{CLICK(4)} \\
	& \texttt{SCROLL(DOWN)} \\
	& \texttt{GOBACK()} \\
	& \texttt{{`}{`}{`}} \\
	&  \\
	& You need to first think about the reasoning process as an internal monologue and then provide the user with an action. Respond in the following format: \texttt{<think>} \\
	& ... \\
	& \texttt{</think>} \\
	& \texttt{<action>} \\
	& ... \\
	& \texttt{</action>}. For example: \\
	&  \\
	& \texttt{<think>} \\
	& I need to have a thinking before I take my action. \\
	& \texttt{</think>} \\
	& \texttt{<action>} \\
	& \texttt{ANSWER("I can take any available action, e.g., give an answer.")} \\
	& \texttt{</action>} \\
	&  \\
	& Note that all the examples above are just examples demonstrating action usage and output format and takes NO ANY RELATION with the following inputs. Now, take your task. \\
	\textcolor{usercolor}{User: }
	& Completed instructions: \\
	& \texttt{\$\{history\_instructions\}} \\
	&  \\
	& Current instruction: \\
	& \texttt{\$\{instruction\}} \\
	&  \\
	& Current Screen: \\
	& \texttt{\$\{screen\}} \\
	& \\
	& Action History: \\
	& \texttt{\$\{action\_history\}} \\
	&  \\
\end{longtable}

\section{Limitations}
\label{sec:limit}

\todo[inline,disable]{1. Performance of LCS labels. 2. More benchmarks.}

Although the proposed \progrm{} achieves the best results in our experiments, we find that the effectiveness of
the LCS-based progress label still holds a remarkable gap with that of the environment-reward-based progress label,
both in the performance of the resulting actor and the progress estimation error. There are still many aspects to polish 
in the current LCS-based progress labeling algorithm, such as soft match function $f$ design (see \cref{sub:slcs_def}),
garbage action (\textit{e.g.}, meaningless empty or scrolling actions) cleaning, \textit{etc}. 

Current experiment results have demonstrated the promising effectiveness of \progrm{} in GUI agent training. The
selected environment also gives the opportunity to obtain a deeper insight into the proposed LCS-based progress labeling
algorithm by comparing it with environment-reward-based progress labeling, which is not supported in other environments.
However, it can still be doubted if \progrm{} can still work well in other GUI environments. Besides, \progrm{}
will be most valuable to be applied to ethe nvironment without well-annotated RL training tasks, as it is expected to efficiently
and accurately evaluate the LLM-generated task goals and alleviate the scarcity of well-annotated RL training tasks.

\section{Broader Societal Impacts}
\label{sec:brd_impct}

The proposed \progrm{} can be used to train more capable GUI agents, which may bring significant convenience to human users by automating a wide range of tasks in GUI systems. However, alongside these benefits, there are potential risks. More powerful GUI agents could be misused to bypass CAPTCHAs, gain unauthorized access to public internet systems, or perform other malicious activities. Additionally, as GUI agents are not yet perfectly reliable, there is a risk that unexpected or dangerous actions could be taken, potentially causing irreparable damage to data or systems.

Overall, the broader societal impacts of \progrm{} are primarily realized through the downstream use of the trained GUI agents, rather than from \progrm{} itself. Responsible deployment and careful consideration of security and safety are therefore essential.

\end{document}